\documentclass[runningheads,a4paper]{llncs}

\usepackage{amssymb}
\usepackage{graphicx}
\usepackage{amsmath}
\usepackage{caption} 
\usepackage{subcaption}
\usepackage{array}
\newcolumntype{M}[1]{>{\raggedright\arraybackslash}m{#1}}

\usepackage{url}
\urldef{\mailsa}\path|{alfred.hofmann, ursula.barth, ingrid.haas, frank.holzwarth,|
\urldef{\mailsb}\path|anna.kramer, leonie.kunz, christine.reiss, nicole.sator,|
\urldef{\mailsc}\path|erika.siebert-cole, peter.strasser, lncs}@springer.com|

\begin{document}

\mainmatter  

\title{Myocardial Segmentation of Contrast Echocardiograms Using Random Forests\\
Guided by Shape Model}

\titlerunning{Myocardial Segmentation Using Shape Model Guided RF}

%
%
\author{Yuanwei Li\inst{1}\and Chin Pang Ho\inst{2}\and Navtej Chahal\inst{3}\and\\
Roxy Senior\inst{3}\and Meng-Xing Tang\inst{1}}
%
\authorrunning{Y. Li et al.}

\institute{Department of Bioengineering, Imperial College London, UK
\and
Department of Computing, Imperial College London, UK
\and
Department of Echocardiography, Royal Brompton Hospital, London, UK
}

%
%

\toctitle{Lecture Notes in Computer Science}
\tocauthor{Authors' Instructions}
\maketitle

\begin{abstract}
Myocardial Contrast Echocardiography (MCE) with micro-bubble contrast agent enables myocardial perfusion quantification which is invaluable for the early detection of coronary artery diseases. In this paper, we proposed a new segmentation method called Shape Model guided Random Forests (SMRF) for the analysis of MCE data. The proposed method utilizes a statistical shape model of the myocardium to guide the Random Forest (RF) segmentation in two ways. First, we introduce a novel Shape Model (SM) feature which captures the global structure and shape of the myocardium to produce a more accurate RF probability map. Second, the shape model is fitted to the RF probability map to further refine and constrain the final segmentation to plausible myocardial shapes. Evaluated on clinical MCE images from 15 patients, our method obtained promising results (Dice=0.81, Jaccard=0.70, MAD=1.68 mm, HD=6.53 mm) and showed a notable improvement in segmentation accuracy over the classic RF and its variants.
\end{abstract}
\section{Introduction}
Myocardial Contrast Echocardiography (MCE) is a cardiac ultrasound imaging technique that utilizes vessel-bound microbubbles as contrast agents. In contrast to conventional B-mode echocardiography which only captures the structure and motion of the heart, MCE also allows for the assessment of myocardial perfusion through the controlled destruction and replenishment of microbubbles \cite{wei_quantification_1998}. The additional perfusion information gives it great potential for the detection of coronary artery diseases. However, current perfusion analysis of MCE data mainly relies on human visual assessment which is time consuming and not reproducible \cite{DBLP:conf/fimh/MaSRBL09}. There is generally a lack of automatic computerized algorithms and methods to help clinician perform accurate perfusion quantification \cite{DBLP:conf/fimh/MaSRBL09}. One major challenge is the automatic segmentation of the myocardium before subsequent perfusion analysis can be carried out.

In this paper, we extend the Random Forests (RF) framework \cite{DBLP:journals/ml/Breiman01} to segment the myocardium in our MCE data. RF is a machine learning technique that has gained increasing use in the medical imaging field for tasks such as segmentation \cite{lempitsky_random_2009} and organ localization \cite{DBLP:journals/mia/CriminisiRKSPWS13}. RF has been successful due to its accuracy and computational efficiency. Promising results of myocardial delineation on 3D B-mode echo has also been demonstrated in \cite{lempitsky_random_2009}. However, classic RF has two limitations. First, our MCE data exhibit large sources of intensity variations \cite{tang_quantitative_2011} due to factors such as speckle noise, low signal-to-noise ratio, attenuation artefacts, unclear and missing myocardial borders, presence of structures (papillary muscle) with similar appearance to the myocardium. These intensity variations reduce the discriminative power of the classic RF that utilizes only local intensity features. Second, RF segmentation operates on a pixel basis where the RF classifier predicts a class label for each pixel independently. Structural relationships and contextual dependencies between pixel labels are ignored \cite{DBLP:conf/iccv/KontschiederBBP11,DBLP:conf/ipmi/MontilloSWIMC11} which results in segmentation with inconsistent pixel labelling leading to unsmooth boundaries, false detections in the background and holes in the region of interest. To overcome the above two problems, we need to incorporate prior knowledge of the shape of the structure and use additional contextual and structural information to guide the RF segmentation.
\\
\indent
There are several works which have incorporated local contextual information into the RF framework. Lempitsky et al. \cite{lempitsky_random_2009} use the pixel coordinates as position features for the RF so that the RF learns the myocardial shape implicitly. Tu et Bai \cite{DBLP:journals/pami/TuB10} introduce the concept of auto-context which can be applied to RF by using the probability map predicted by one RF as features for training a new RF. Montillo et al. \cite{DBLP:conf/ipmi/MontilloSWIMC11} extend the auto-context RF by introducing entanglement features that use intermediate probabilities derived from higher levels of a tree to train its deeper levels. Kontschieder et al. \cite{DBLP:conf/iccv/KontschiederBBP11} introduce the structured RF that builds in structural information by using RF that predicts structured class labels for a patch rather than the class label of an individual pixel. Lombaert et al. \cite{DBLP:conf/miccai/LombaertCA15} use spectral representations of shapes to classify surface data.
\\
\indent
The above works use local contextual information that describes the shape of a structure implicitly. The imposed structural constraint are not strong enough to guide the RF segmentation in noisy regions of MCE data. In this paper, we proposed the Shape Model guided Random Forests (SMRF) which provides a new way to incorporate global contextual information into the RF framework by using a statistical shape model that captures the explicit shape of the entire myocardium. This imposes stronger, more meaningful structural constraints that guide the RF segmentation more accurately. The shape model is learned from a set of training shapes using Principal Component Analysis (PCA) and is originally employed in Active Shape Model (ASM) where the model is constrained so that it can only deform to fit the data in ways similar to the training shapes \cite{DBLP:journals/cviu/CootesTCG95}. However, ASM requires a manual initialization and the final result is sensitive to the position of this initialization. Our SMRF is fully automatic and enjoys both the local discriminative power of the RF and the prior knowledge of global structural information contained in the statistical shape model. The SMRF uses the shape model to guide the RF segmentation in two ways. First, it directly incorporates the shape model into the RF framework by introducing a novel Shape Model (SM) feature which has outperformed the other contextual features and produced a more accurate RF probability map. Second, the shape model is fitted to the probability map to generate a smooth and plausible myocardial boundary that can be used directly for subsequent perfusion analysis.

\section{Method}
In this section, we first review some basic background on statistical shape model and RF. We then introduce the two key aspects of our SMRF---the novel SM feature and the fitting of the shape model.
\subsubsection{\textit{Statistical Shape Model:}}
\begin{figure}
\centering
\begin{subfigure}[b]{.5\textwidth}
  \centering
  \includegraphics[height=0.16\textheight]{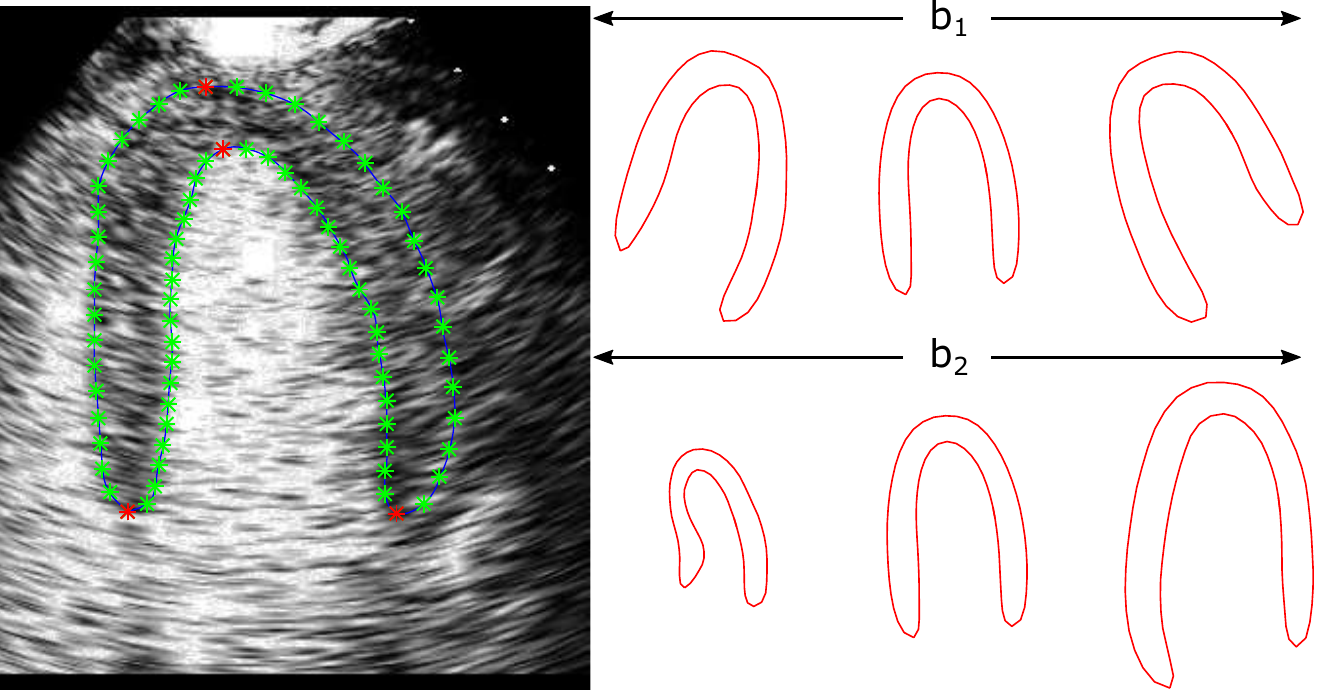}
  \caption{}
  \label{fig:Model}
\end{subfigure}%
\begin{subfigure}[b]{.5\textwidth}
  \centering
  \includegraphics[height=0.16\textheight]{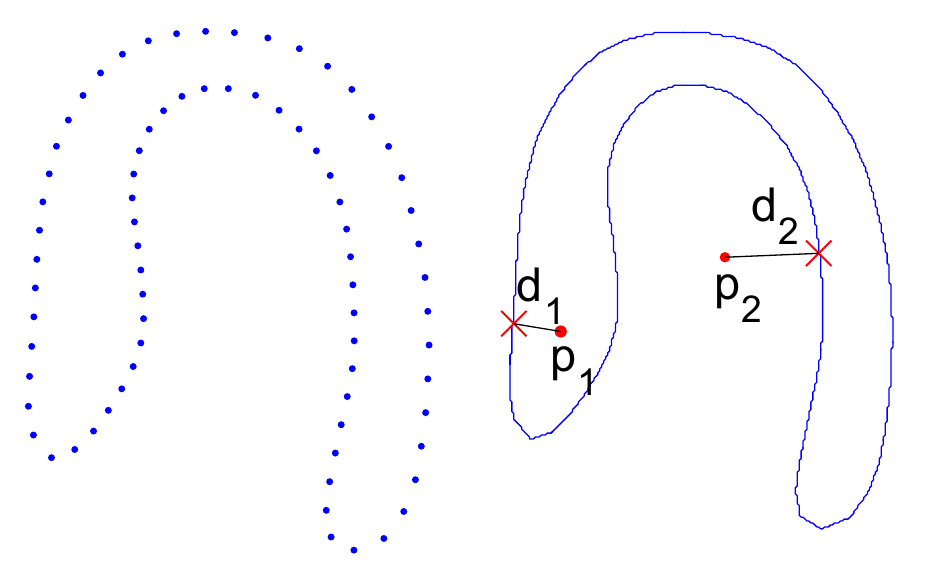}
  \caption{}
  \label{fig:Feature}
\end{subfigure}
\caption{(a) Left: A manual annotation from training set showing key landmarks (\textit{red}) and other landmarks in between (\textit{green}). Right: First two modes of variations of the shape model. (b) Left: Landmarks $\boldsymbol{x}$ (\textit{blue dots}) generated randomly by the shape model in (\ref{eq:Model}). Right: $d_1$($d_2$) is the SM feature value measuring the signed shortest distance from pixel $\boldsymbol{p_1}$($\boldsymbol{p_2}$) to the myocardial boundary $B(\boldsymbol{x})$ (\textit{blue contour}). $d_1$ is positive and $d_2$ is negative.}
\label{fig:Result}
\end{figure}

A statistical shape model of the myocardium is built from 89 manual annotations using PCA \cite{DBLP:journals/cviu/CootesTCG95}. Each annotation has $N=76$ landmarks comprising 4 key landmarks with 18 landmarks spaced equally in between along the boundary of manual tracing (Fig. \ref{fig:Model} left). The shape model is represented as:
\begin{equation}\label{eq:Model}
  \boldsymbol{x}=\bar { \boldsymbol{x} } + \boldsymbol{Pb}
\end{equation}
\noindent where $\boldsymbol{x}$ is a 2$N$-D vector containing the 2D coordinates of the $N$ landmark points, $\bar{\boldsymbol{x}}$ is the mean coordinates of all training shapes, $\boldsymbol{b}$ is a set of $K$ shape parameters and $\boldsymbol{P}$ contains $K$ eigenvectors with their associated eigenvalues ${ \lambda  }_{ i }$. $K$ is the number of modes and set to 10 to explain 98\% of total variance so that fine shape variations are modeled while noise is removed. Values of ${b}_{i}$ are bounded between $\pm s\sqrt { \lambda _{ i } } $ so that only plausible shape similar to the training set is generated (Fig. \ref{fig:Model} right). Refer to \cite{DBLP:journals/cviu/CootesTCG95} for details on statistical shape model.
\subsubsection{\textit{Random Forests:}}
Myocardial segmentation can be formulated as a problem of binary classification of image pixels. An RF classifier \cite{DBLP:journals/ml/Breiman01} is developed that predicts the class label (myocardium or background) of a pixel using a set of features. The RF is an ensemble of decision trees. During training, each branch node of a tree learns a pair of feature and threshold that results in the best split of the training pixels into its child nodes. The splitting continues recursively until the maximum tree depth is reached or the number of training pixels in the node falls below a minimum. At this time, a leaf node is created and the class distribution of the training pixels reaching the leaf node is used to predict the class label of unseen test pixels. The average of the predictions from all the trees gives a segmentation probability map. Refer to \cite{DBLP:journals/ml/Breiman01}, \cite{lempitsky_random_2009} for details on RF.
\subsubsection{\textit{Shape Model Feature:}}
The classic RF uses local appearance features which are based on surrounding image intensities of the reference pixel \cite{DBLP:journals/mia/CriminisiRKSPWS13}. We introduced an additional novel SM feature that is derived from the shape model. The SM feature randomly selects some values for the shape model parameters $\boldsymbol{b}$ and generates a set of landmarks $\boldsymbol{x}$ using (\ref{eq:Model}) (Fig. \ref{fig:Feature} left). The landmarks can be joined to form a myocardial boundary. Let $B(\bar { \boldsymbol{x} } +\boldsymbol{Pb})$ be the myocardial boundary formed by joining the landmarks generated using some values of $\boldsymbol{b}$. The SM feature value is then given by the signed shortest distance $d$ from the reference pixel $\boldsymbol{p}$ to the boundary $B$ (Fig. \ref{fig:Feature} right). The distance is positive if $\boldsymbol{p}$ lies inside the boundary and negative if it lies outside. The SM feature is essentially the signed distance transform of a myocardial boundary generated by the shape model. Each SM feature is defined by the shape parameters $\boldsymbol{b}$. During training, an SM feature is created by random uniform sampling of each $b_{i}$ in the range of $\pm s_{feature}\sqrt { \lambda _{ i } }$ where $s_{feature}$ is set to 1 in all our experiments. The binary SM feature test, parameterized by $\boldsymbol{b}$ and a threshold $\tau$, is written as:
\begin{equation}
{ t }_{ \textrm{SM} }^{ \boldsymbol{b},\tau  }(\boldsymbol{p})=\begin{cases} 1,\qquad \textrm{if}\quad D(\boldsymbol{p},B(\bar { \boldsymbol{x} } +\boldsymbol{Pb}))>\tau  \\ 0,\qquad \textrm{otherwise}. \end{cases}
\end{equation}
\noindent where $D(.)$ is the function that computes $d$. Depending on the binary test outcome, pixel $\boldsymbol{p}$ will go to the left (1) or right (0) child node of the current split node. During training, the RF learns the values of $\boldsymbol{b}$ and $\tau$ that best split the training pixels at a node. The SM features explicitly impose a global shape constraint in the RF framework. The random sampling of $\boldsymbol{b}$ also allows the RF to learn plausible shape variations of the myocardium.
\subsubsection{\textit{Shape Model Fitting:}}
The RF output is a probability map which cannot be used directly in subsequent analysis and application. Simple post-processing on the probability map such as thresholding and edge detection can produce segmentations with inaccurate and incoherent boundaries due to the nature of the pixel-based RF classifier. Our SMRF fits the shape model to the RF probability map to extract a final myocardial boundary that is smooth and which preserves the integrity of the myocardial shape. The segmentation accuracy is also improved as the shape constraint imposed by the shape model can correct some of the misclassifications made by the RF.

Let ${ T }_{ \boldsymbol{\theta}}$ be a pose transformation defined by the pose parameter $\boldsymbol{\theta}$ which includes translation, rotation and scaling. The shape model fitting is then an optimization problem where we want to find the optimal values of $(\boldsymbol{b},\boldsymbol{\theta})$ such that the model best matches the RF probability map under some shape constraints. We minimize the following objective function:
\begin{equation}
\begin{aligned}
& \underset{\boldsymbol{b},\boldsymbol{\theta}}{\text{min}}
& & { { \left\| { \boldsymbol{I} }_{ \textrm{RF} }-{ \boldsymbol{I} }_{ \textrm{M} }({ T }_{ \boldsymbol{\theta}  }(\bar { \boldsymbol{x} } +\boldsymbol{Pb})) \right\|  }^{ 2 }+\alpha \frac { 1 }{ K } \sum _{ i=1 }^{ K }{ \frac { \left| { b }_{ i } \right|  }{ \sqrt { { \lambda  }_{ i } }  }  }  } \\
& \text{subject to}
& & -{ s }_{ fit }\sqrt { { \lambda  }_{ i } } <{ b }_{ i }<{ s }_{ fit }\sqrt { { \lambda  }_{ i } }, \; i = 1, \ldots, K.
\end{aligned}
\end{equation}
The first term of the objective function compares how well the match is between the model and the RF probability map $\boldsymbol{I}_{\textrm{RF}}$. $\boldsymbol{I}_{\textrm{M}}(.)$ is a function that converts the landmarks generated by the shape model into a binary mask of the myocardial shape. This allows us to evaluate a dissimilarity measure between the RF segmentation and the model by computing the sum of squared difference between the RF probability map and the model binary mask. The second term of the objective function is a regularizer which imposes a shape constraint. It is related to the probability of a given shape \cite{DBLP:journals/pr/CristinacceC08} and ensures that it does not deviate too much away from its mean shape. $\alpha$ is the weighting given to the regularization term and its value is determined empirically. Finally, an additional shape constraint is imposed on the objective function by limiting the upper and lower bounds of $b_i$ to allow for only plausible shapes. $s_{fit}$ is set to 2 in all our experiments. The optimization is carried out using direct search which is a derivative-free solver from the MATLAB global optimization toolbox. At the start of the optimization, each $b_i$ is initialized to zero. Pose parameters are initialized such that the model shape is positioned in the image center with no rotation and scaling.
\section{Experiments}
\subsubsection{\textit{Datasets:}}
2D+t MCE sequences were acquired from 15 individuals using a Philips iE33 ultrasound machine and SonoVue as the contrast agent. Each sequence is taken in the apical 4-chamber view under the triggered mode which shows the left ventricle at end-systole. One 2D image was chosen from each sequence and the myocardium manually segmented by two experts to give inter-observer variability. This forms a dataset of 15 2D MCE images for evaluation. Since the appearance features of the RF are not intensity invariant, all the images are pre-processed with histogram equalization to reduce intensity variations between different images. The image size is approximately 351$\times$303 pixels.
\begin{figure}
\centering
\begin{subfigure}[b]{.55\textwidth}
  \centering
  \includegraphics[width=\linewidth]{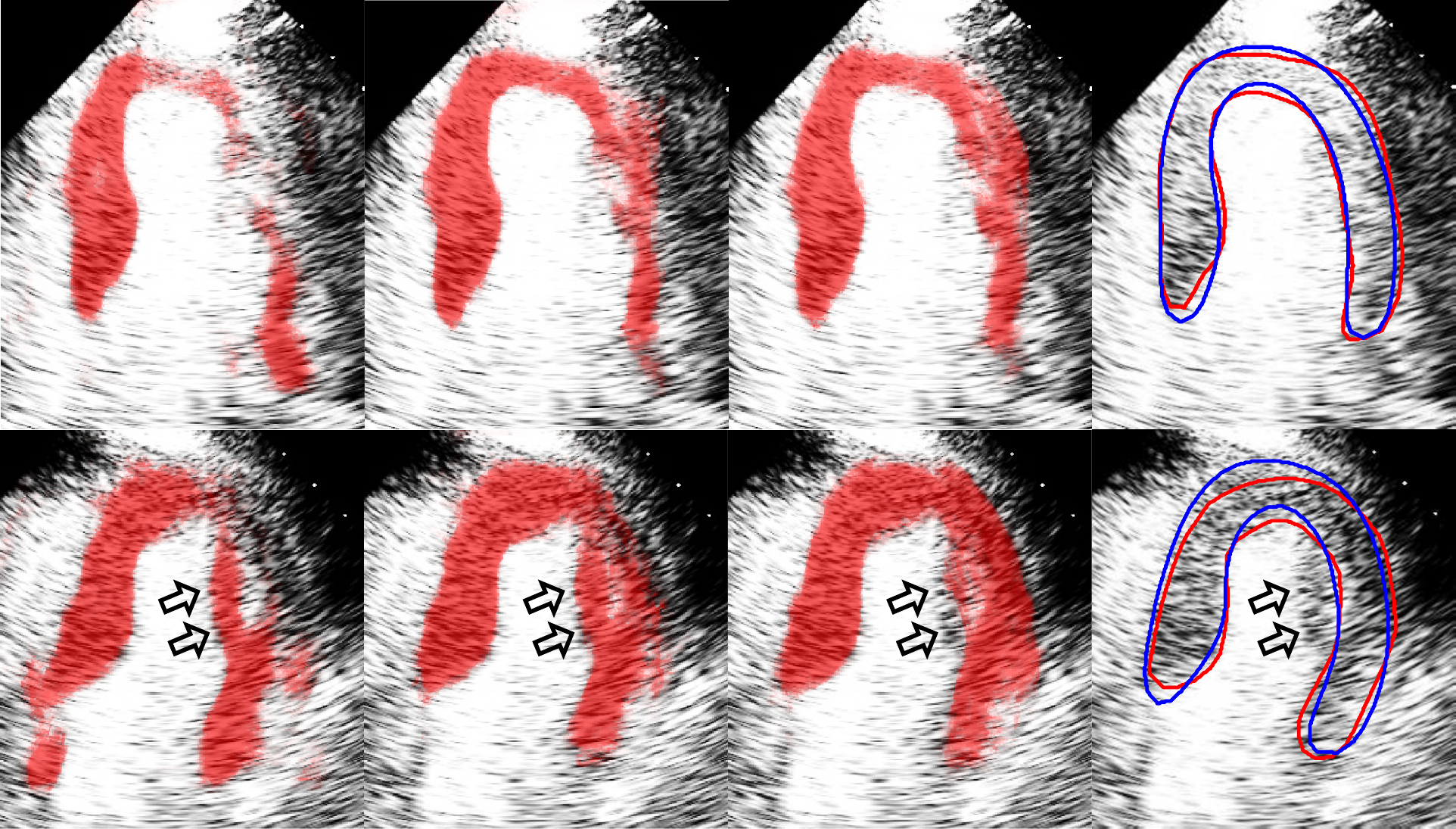}
  \caption{}
  \label{fig:ResultVisual}
\end{subfigure}%
\begin{subfigure}[b]{.45\textwidth}
  \centering
  \includegraphics[width=\linewidth]{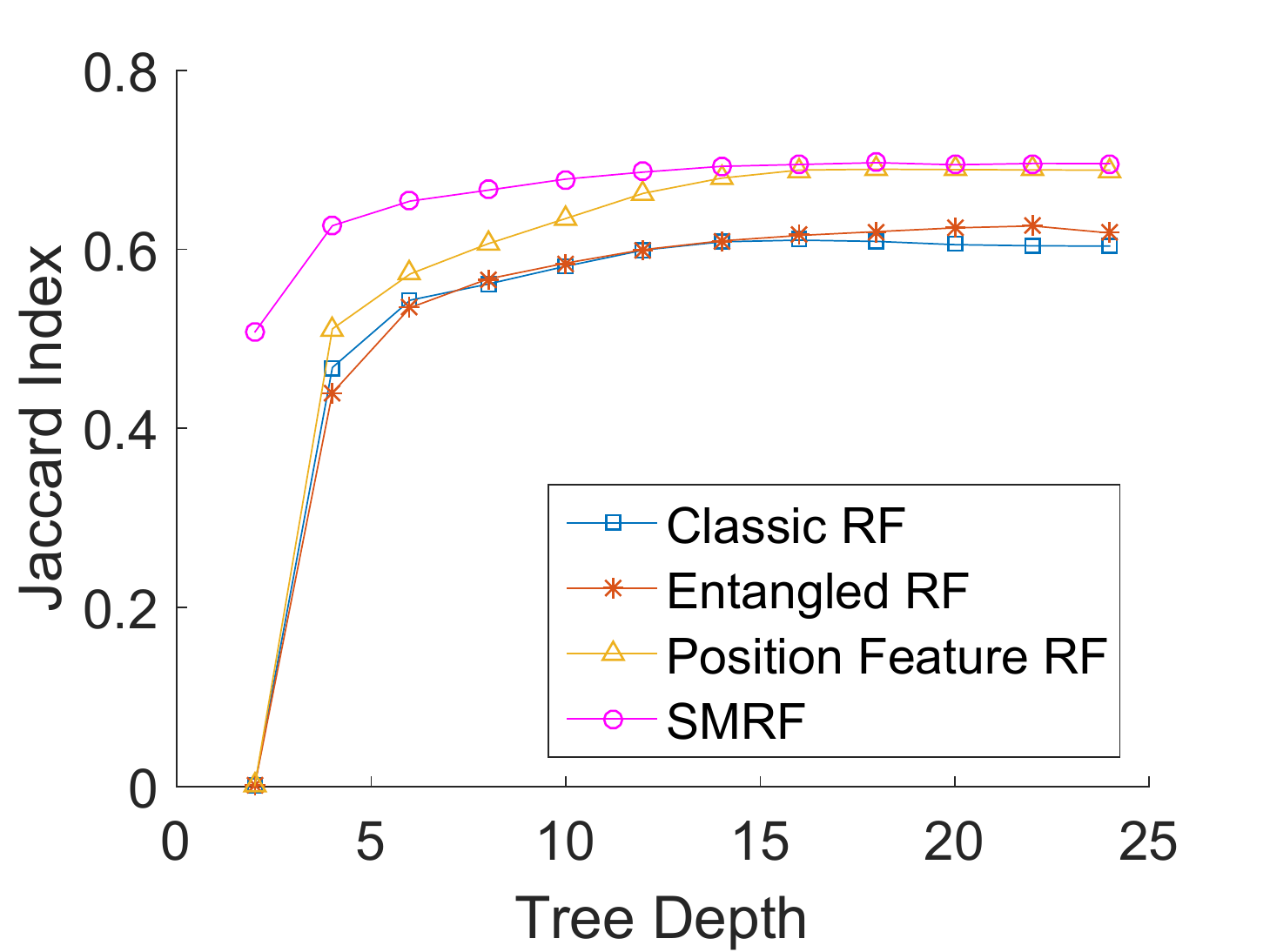}
  \caption{}
  \label{fig:ResultQuant}
\end{subfigure}
\caption{(a) Visual segmentation results with one MCE example on each row. First three columns: Probability maps from classic RF, position feature RF and SMRF respectively. Last column: Ground truth boundary (\textit{red}) and the SMRF boundary (\textit{blue}) obtained from fitting the shape model to the SMRF probability map in the third column. \textit{Black arrows} indicate papillary muscle. (b) Segmentation accuracy of different RF classifiers at different tree depths.}
\label{fig:Result}
\end{figure}
\subsubsection{\textit{Validation Methodology:}}
We compared our SMRF segmentation results to the classic RF that uses appearance features \cite{DBLP:journals/mia/CriminisiRKSPWS13}, as well as RFs that use other contextual features such as entanglement \cite{DBLP:conf/ipmi/MontilloSWIMC11} and position features \cite{lempitsky_random_2009}. We also compared our results to repeated manual segmentations and the Active Shape Model (ASM) method \cite{DBLP:journals/tmi/GinnekenFSRV02}. Segmentation accuracy is assessed quantitatively using pixel classification accuracy, Dice and Jaccard indices, Mean Absolute Distance (MAD) and Hausdorff Distance (HD). To compute the distance error metrics (MAD and HD), a myocardial boundary is extracted from the RF probability map using the Canny edge detector. This is not required for the SMRF in which the shape model fitting step directly outputs a myocardial boundary.
\\
\indent
We performed leave-one-out cross-validation on our dataset of 15 images. The RF parameters are determined experimentally and then fixed for all experiments. 20 trees are trained with maximum tree depth of 24. 10\% of the pixels from the training images are randomly selected for training. The RF and the shape model fitting were implemented in C\# and MATLAB respectively. Given an unseen test image, RF segmentation took 1.5min with 20 trees and shape model fitting took 8s on a machine with 4 cores and 32GB RAM. RF training took 38mins.
\section{Results}
Fig. \ref{fig:ResultVisual} qualitatively shows that our SMRF probability map (column 3) has smoother boundary and more coherent shape than the classic RF (column 1) and position feature RF (column 2). Fitting the shape model to the SMRF probability map produces the myocardial boundary (\textit{blue}) in column 4. The fitting guides the RF segmentation especially in areas where the probability map has a low confidence prediction. In the example on the second row, our SMRF predicts a boundary that correctly excludes the papillary muscle (\textit{black arrows}). This is often incorrectly included by the other RFs due to its similar appearance to the myocardium.
\begin{table}[]
\centering
\caption{Quantitative comparison of segmentation results between the proposed SMRF and other methods. Results presented as (Mean $\pm$ Standard Deviation).}
\label{table:Result}
\begin{tabular}{|M{10.5em}|c|c|c|c|r|}
\hline
 & \multicolumn{1}{c|}{Accuracy} & \multicolumn{1}{c|}{Dice} & \multicolumn{1}{c|}{Jaccard} & \multicolumn{1}{c|}{\begin{tabular}[c]{@{}c@{}}MAD\\   (mm)\end{tabular}} & \multicolumn{1}{c|}{\begin{tabular}[c]{@{}c@{}}HD\\   (mm)\end{tabular}} \\ \hline
Intra-observer & 0.96$\pm$0.01 & 0.89$\pm$0.02 & 0.80$\pm$0.03 & 1.02$\pm$0.26 & 3.75$\pm$0.93 \\
Inter-observer & 0.94$\pm$0.02 & 0.84$\pm$0.05 & 0.72$\pm$0.07 & 1.59$\pm$0.57 & 6.90$\pm$3.24 \\
ASM \cite{DBLP:journals/tmi/GinnekenFSRV02} & 0.92$\pm$0.03 & 0.77$\pm$0.08 & 0.64$\pm$0.11 & 2.23$\pm$0.81 & 11.44$\pm$5.23 \\ \hline
Classic RF & 0.91$\pm$0.04 & 0.74$\pm$0.12 & 0.60$\pm$0.14 & 2.46$\pm$1.36 & 15.69$\pm$7.34 \\
Entangled RF \cite{DBLP:conf/ipmi/MontilloSWIMC11} & 0.91$\pm$0.05 & 0.75$\pm$0.13 & 0.62$\pm$0.15 & 2.43$\pm$1.62 & 15.06$\pm$7.92 \\
Position Feature RF \cite{lempitsky_random_2009} & \textbf{0.93$\pm$0.03} & \textbf{0.81$\pm$0.10} & 0.69$\pm$0.13 & 1.81$\pm$0.84 & 9.51$\pm$3.80 \\
SMRF & \textbf{0.93$\pm$0.03} & \textbf{0.81$\pm$0.10} & \textbf{0.70$\pm$0.12} & \textbf{1.68$\pm$0.72} & \textbf{6.53$\pm$2.61} \\ \hline
\end{tabular}
\end{table}

Table. \ref{table:Result} compares the quantitative segmentation results of our SMRF to other methods. Both SM features and position features encode useful structural information that produces more accurate RF probability maps than the classic RF and entangled RF. This is reflected by the higher Dice and Jaccard indices. For MAD and HD metrics, SMRF outperforms all other RF methods because the shape model fitting step in SMRF produces more accurate myocardial boundaries than those extracted using the Canny edge detector. In addition, SMRF also outperforms ASM \cite{DBLP:journals/tmi/GinnekenFSRV02} and comes close to the inter-observer variations.

Fig. \ref{fig:ResultQuant} compares the segmentation accuracy of the probability maps of different RF classifiers. Our SMRF obtained higher Jaccard indices than the classic and entangled RFs at all tree depths. At lower tree depths, SMRF shows notable improvement over the position feature RF. The SM features have more discriminative power than the position features as it captures the explicit geometry of the myocardium using the shape model. The SM feature binary test partitions the image space using more complex and meaningful myocardial shapes as opposed to position feature which simply partitions the image space using straight lines. This provides a stronger global shape constraint than the position feature and allows a decision tree to converge faster to the correct segmentation at lower tree depths. This gives the advantage of using trees with smaller depths which speeds up both training and testing.
\section{Conclusion}
We presented a new method SMRF for myocardial segmentation in MCE images. We showed how our SMRF utilizes a statistical shape model to guide the RF segmentation. This is particular useful for MCE data whose image intensities are affected by many variables and therefore prior knowledge of myocardial shape becomes important in guiding the segmentation. Our SMRF introduces a new SM feature which captures the global myocardial structure. This feature outperforms other contextual features to allow the RF to produce a more accurate probability map. Our SMRF then fits the shape model to the RF probability map to produce a smooth and coherent final myocardial boundary that can be used in subsequent perfusion analysis. In future work, we plan to validate our SMRF on a larger, more challenging dataset which includes different cardiac phases and chamber views.
\subsubsection*{Acknowledgments.} The authors would like to thank Prof. Daniel Rueckert, Liang Chen and other members from the BioMedIA group for their help and advice. This work was supported by the Imperial College PhD Scholarship.
\bibliography{References}
\end{document}